**Pedram Ghamisi**   p.ghamisi@gmail.com

Remote Sensing Technology Institute (IMF), German Aerospace Center (DLR), Germany.
Signal Processing in Earth Observation (SiPEO), Technische Universität München (TUM), Germany.

**Gabriele Cavallaro**   cavallaro.gabriele@gmail.com

Faculty of Electrical and Computer Engineering, University of Iceland, 107 Reykjavik, Iceland.

**Dan (Sabrina) Wu**   cavallaro.gabriele@gmail.com

Centre for Spatial Environmental Research, School of Geography, Planning and Environmental Management, University of Queensland, Australia

**Jon Atli Benediktsson**   benedikt@hi.is

Faculty of Electrical and Computer Engineering, University of Iceland, 107 Reykjavik, Iceland.

**Antonio Plaza**   aplaza@unex.es

Department of Technology of Computers and Communications, University of Extremadura, Spain.

This research has been supported by the Icelandic Research Fund for Graduate Students and Alexander von Humboldt Fellowship for postdoctoral researchers.

# Integration of LiDAR and Hyperspectral Data for Land-cover Classification: A Case Study

**Abstract:** In this paper, an approach is proposed to fuse LiDAR and hyperspectral data, which considers both spectral and spatial information in a single framework. Here, an extended self-dual attribute profile (ESDAP) is investigated to extract spatial information from a hyperspectral data set. To extract spectral information, a few well-known classifiers have been used such as support vector machines (SVMs), random forests (RFs), and artificial neural networks (ANNs). The proposed method accurately classify the relatively volumetric data set in a few CPU processing time in a real ill-posed situation where there is no balance between the number of training samples and the number of features. The classification part of the proposed approach is fully-automatic.

**Keywords:** Hyperspectral; Light Detection and Ranging (LiDAR); self-dual attribute profiles (SDAPs); fusion of hyperspectral and LiDAR data.

I. INTRODUCTION

LiDAR data have been investigated for different applications in urban, rural and forested areas such as building classification [1], roof plane detection [2], vegetation classification [3], quantifying riparian habitat structure [4], and classification of complex forest areas [5]. However, the use of LiDAR data for complex areas (e.g., where many classes are located close to each other) is very limited compared to optical data (e.g., multispectral and hyperspectral data) due to the lack of spectral information provided by this type of sensors [6]. To address this shortcoming, LiDAR technology can be jointly used along with optical sensors [7]. In this way, some space-borne broadband multispectral sensors are capable of capturing data with a very high spatial resolution (e.g., GeoEye and IKONOS) but their spectral resolution is very limited which downgrades their capability for discriminating different classes in complex areas. Hyperspectral sensors have proven to be efficient for discriminating complex materials due to their high spectral resolution [8]. Goodenough et al. in [9] compared classification results of a forested area captured by three different sensors, two multispectral sensors (i.e., Landsat-7 ETM+ and EO-1 ALI), and one hyperspectral sensor (EO-1 Hyperion). The obtained results confirmed that the hyperspectral observations can significantly improve the result of the multispectral measurements in terms of classification accuracies. However, hyperspectral data do not provide any information about the elevation and size of different materials. since, for example, urban areas are composed of a variety of objects which are made up of similar materials with almost the same spectral characteristics (e.g., road and roof), hyperspectral data alone may not able to efficiently discriminate such materials.

As mentioned in [10], urban environments are highly complex and challenging and it is optimistic to assume that a single sensor can provide all the necessary information that one may demand for object characterization. Due to the considerable increase in the number of available instruments for Earth observation in recent years, it is now possible to have different types of data acquired over the same scene on the Earth by different sensors. This increase has led to many research works related to the joint use of passive and active sensors for the accurate classification of different materials. In this line of thought, hyperspectral data can provide detailed spectral information, which makes the analysis and classification of complex areas possible. LiDAR data can provide in turn useful information regarding the size and elevation of different objects. As a result, by emphasizing on the strengths of each sensor, an integration of the two could potentially lead to an improvement in classification results.

The joint use of hyperspectral and LiDAR data has proven to be successful for a wide variety of applications such as shadow, height, and gap related masking techniques [11–13], above-ground biomass estimates [14], quantifying riparian habitat structure [15], and fuel type mapping [16]. In addition, the joint use of LiDAR and hyperspectral data has led to higher classification accuracies compared to the use of each one separately. For instance, in [5, 17–20], spatial and structural information obtained by LiDAR data have been used in addition to spectral information from multispectral and hyperspectral sensors and an improved discrimination of different classes in forested and urban areas was obtained. With respect to the aforementioned works, it can be concluded that LiDAR and hyperspectral data may complement each other well and by integrating those two data sets in a proper way, one can make the most of the advantages of the two, while addressing the shortcomings of each of them. However, as mentioned in [5], most of the studies do not consider the integration of LiDAR and hyperspectral from a real data fusion perspective but use them by considering those sources of information separately. In addition, most of the existing methods are not automatic and their performance is highly dependent on the initialization step. Therefore, existing techniques demand a significant effort by users to initialize the parameters in a trial and error way and different steps need to be performed manually [5]. This makes the existing techniques very time consuming which is not appropriate for applications in which a rapid response is needed. Moreover, in the literature, spatial information has been often discarded, which is considered as a disadvantage for land-cover classification [20].

In this paper, we address the above-mentioned issues by developing an efficient classification workflow, which exploits the classification of LiDAR and hyperspectral data sets in an automatic way. The main motivation of this paper is that, at this point, it is becoming more common to acquire both LiDAR and hyperspectral data from the same scene. However, these data sets have been independently used in different processing steps, while, as mentioned above, it may be advantageous to consider both the sources in a data fusion framework for discriminating different classes of interests precisely. With respect to the results shown later in this paper, it is easy to derive that the proposed approach is able to accurately classify a high volume of data within a short CPU processing time. The main contributions of the paper are as follows:

1) Development of an efficient classifier for the joint use of hyperspectral and LiDAR data. The proposed approach can specifically (i) handle high dimensional data with very limited number of training samples (ill-posed scenarios) efficiently in terms of classification accuracies and with manageable CPU processing time in an automatic way for the classification part; (ii) exploit the complementary information obtained by LiDAR data with respect to hyperspectral data in order to discriminate between different classes of interest in the studied area.

2) Exploitation of extended self-dual attribute profiles (ESDAPs) [21] to extract valuable spatial information that is investigated to enhance the performance of the proposed technique in terms of classification accuracies. Three well-known classifiers including SVMs, RFs and ANNs are used in this work to perform classification based on the spectral and the spatial information simultaneously.

3) The fusion of LiDAR and hyperspectral data is a vibrant research topic in the remote sensing community. In order to push this field forward, we have made our data set available at http://pedram-ghamisi.com/index sub2.html.

The proposed approach was evaluated using a data set including hyperspectral data with a very high spatial resolution (1 m) and LiDAR data with a density higher than two points per square meter. The results obtained indicate that the proposed approach can lead to very high classification accuracy in a very short CPU processing time when only a limited number of training samples are available. The rest of the paper is organized as follows: Section II discusses the investigated data set. Section III elaborates the problem definition and the preprocessing step in detail. The proposed methodology is described in detail in section IV. Section V is devoted to experimental results and finally, section VI outlines the main concluding remarks.

## II. DATA SET DESCRIPTION

The study area around *Samford Ecological Research Facility* (SERF) is within the Samford valley, which is located at coordinates: $27.38572°S$, $152.877098°E$ (central point of the study area) in south east Queensland, Australia. The *Vegetation Management Act 1999* protects the vegetation on this property as it provides a refuge to native flora and fauna that are under increasing pressure from urbanization.

The hyperspectral image was collected by the *SPECIM AsiaEAGLE II* sensor on the second of February, 2013. This sensor captures 252 spectral channels ranging from $400.7nm$ to $999.2nm$. The spatial resolution of the image was set to $1m$. The size of the dataset is 2082 pixels $\times$ 1606 pixels.

The airborne LiDAR data were acquired by the ALTM Leica ALS50-II sensor in 2009 with a total of 3716157 points in the study area. The average flight height was 1700 meters and the average point density is two points per square meter. The laser pulse wavelength is $1064nm$ with a repetition rate of 126 kHz, an average sample spacing of $0.8m$ and a footprint of $0.34m$. The data were collected up to four returns per pulse and the intensity records were supplied on all pulse returns. The nominal vertical accuracy was $\pm0.15m$ at 1 sigma and the measured vertical accuracy was $\pm0.05m$ at 1 sigma (determined from check points located on open clear ground). The measured horizontal accuracy was $0.31m$ at 1 sigma. Ground LiDAR returns were interpolated and rasterized into a $1m \times 1m$ Digital Elevation Model (DEM) that was provided by the LiDAR contractor was created from the LiDAR ground points and interpolated coastal boundaries. The first returns of the airborne LiDAR data were used to produce the Normalized Digital Surface Model (nDSM) at $1m$ spatial resolution using Las2dem. The $1m$ spatial resolution intensity image was also produced using *Las2dem*. *Las2dem* interpolated the points using Triangulated Irregular Networks (TIN), then the TINs were rasterized into the nDSM and the intensity image with a pixel size of 1m. The intensity image with $1m$ spatial resolution was also produced using *Las2dem*. The LiDAR data were classified into "ground" and "non-ground" by the data contractor using algorithms tailored especially for the project area. In the areas covered by dense vegetation, less laser pulse reaches the ground and fewer ground points were available for DEM and nDSM surfaces interpolation. Therefore the DEM and the

5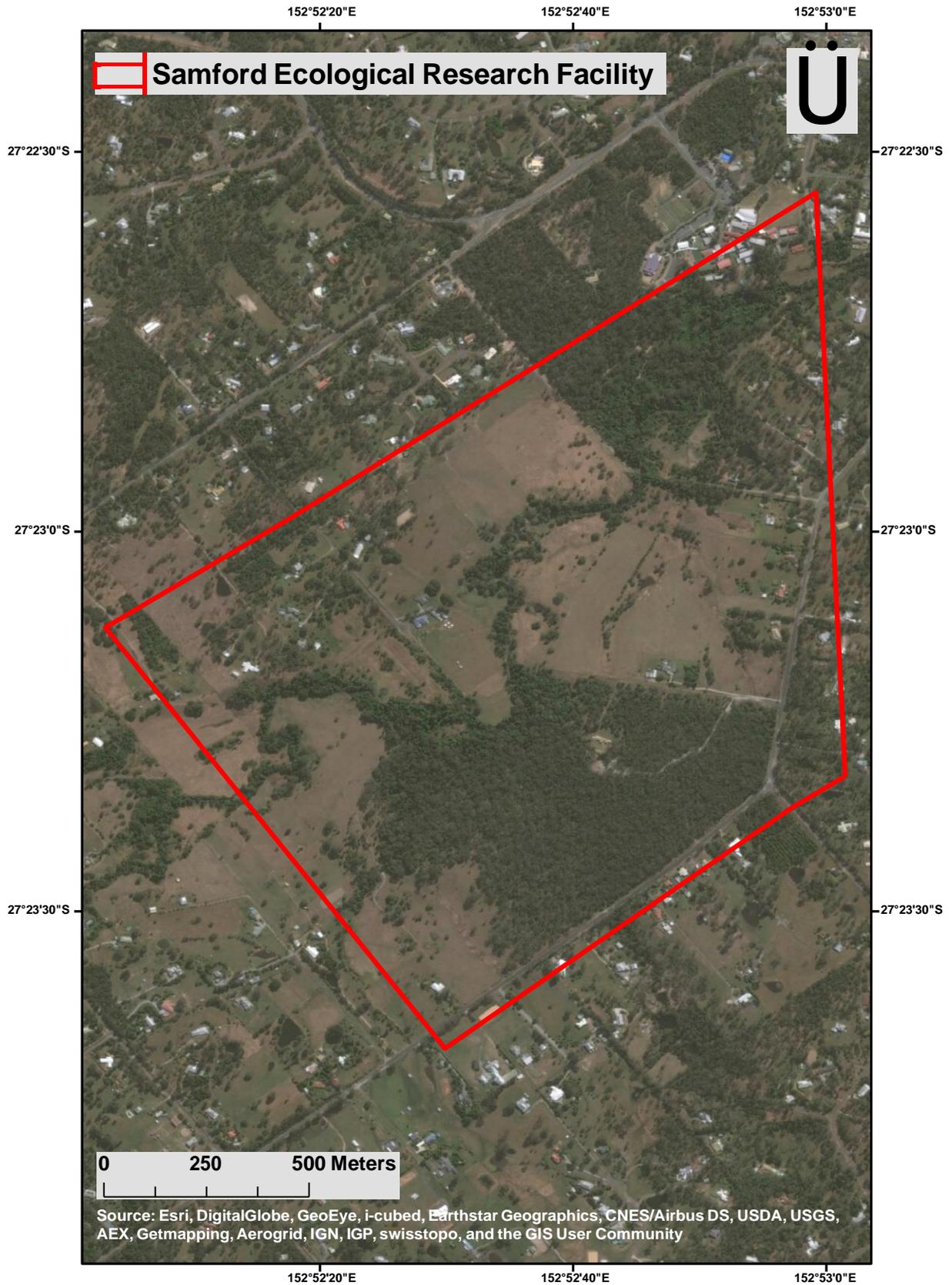

April 8, 2016 DRAFTFig. 1: The study area.

nDSM tend to be less accurate in these areas.

It should be noted that, although hyperspectral and LiDAR data have been captured in 2013 and 2009, respectively, the study area did not have a significant change during the period. Fig. 1 illustrates the study area and Fig. 2 shows band 67 of the hyperspectral image and the obtained nDSM file.

## III. PROBLEM DEFINITION AND PREPROCESSING

Fig. 4 shows the general work flow of the approach developed in this paper. In general, this approach consists of two main phases: 1) preprocessing and 2) classification. The preprocessing step is detailed in this section and the classification approach will be elaborated in the next one. Before describing these techniques, we first provide a definition of the problem.

### A. Problem definition

Hyperspectral imaging sensors usually acquire data with a relatively small field of view. Therefore, a number of partially overlapping images of the investigated area should be integrated according to a mosaic procedure in order to cover a vast area. To do so, let us assume that we have a set of $d$-dimensional hyperspectral images $\mathbf{H}_i$ in which $i = \{1, 2, ..., N\}$. Let us imagine that $\mathbf{H}$ is the final output of the mosaic procedure ($\bigcup \mathbf{H}_i = \mathbf{H}$). Let $\mathbf{L}$ represent the LiDAR derived images including the **nDSM** and the intensity $\mathbf{I}$ (i.e., $\mathbf{L} = \{\mathbf{nDSM}, \mathbf{I}\}$). The goal of our method is to perform the proposed classification approach on the integration of $\mathbf{H}$ and $\mathbf{L}$. The proposed approach performs classification of the data including both $\mathbf{H}$ and $\mathbf{L}$ into $K$ non-overlapping classes $\Omega = \{w_1, w_2, ..., w_K\}$. In this context, the LiDAR and hyperspectral data are jointly classified by using a set of representative samples for each class, referred to as *training samples*. Training samples are usually obtained by manually labeling a small number of pixels in an image or based on some field measurements.

### B. Preprocessing

Six flight lines were used for the hyperspectral data to cover the whole study area. FLAASH (fast line-of-site atmospheric analysis of spectral hypercubes) in ENVI 4.8 was used to atmospherically correct each of the flight lines before mosaicking them into a single image. Fig. 3 shows examples of intensity differences accross the mosaicked lines, which make the classification of this area very challenging due to the dramatic amount of the within-class variances. The Microtops were used to capture solar radiance data in five wavelengths during the airborne hyperspectral imaging campaign, which were used to extract information on aerosol optical thickness for the atmospheric correction. The hyperspectral image was co-registered to the LiDAR intensity image using six ground control points (GCPs) distributed across the image. A polynomial transformation of first order and a nearest-neighbor resampling of the pixels method were applied for the image co-registration. The six GCPs were evenly distributed throughout the study area. The root mean square error computed in this process was 0.643*m*.

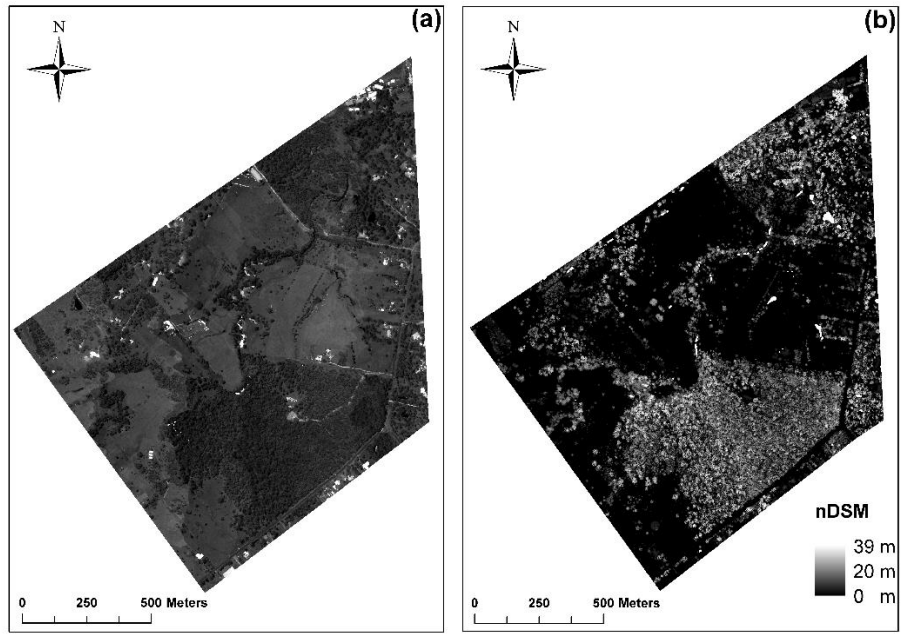

Fig. 2: An illustration of the data set used in experiments including: ( a) band 67 (551.81 nm) of the hyperspectral image, and (b) nDSM file.

## IV. Classification approach

As can be seen from Fig. 4, in order to produce the final classification map, we first apply kernel principal component analysis (KPCA) [22] to the input hyperspectral data and the first informative features with cumulative variance of at least 95% are kept since they contain most of the variance in the data set within a very few features, i.e., only three features are kept as the cumulative sum of eigenvalues of them in percentage is 95.63%. Then, SDAPs are constructed for the retained features in order to extract spatial information. Furthermore, the features obtained by applying the ESDAP on the output of KPCA along with the nDSM and the intensity image are concatenated into a stacked vector. The final classification map is produced by performing SVM [23], RF [24, 25] or radial basis function NN (RBFNN) [26, 27] on the stacked features.

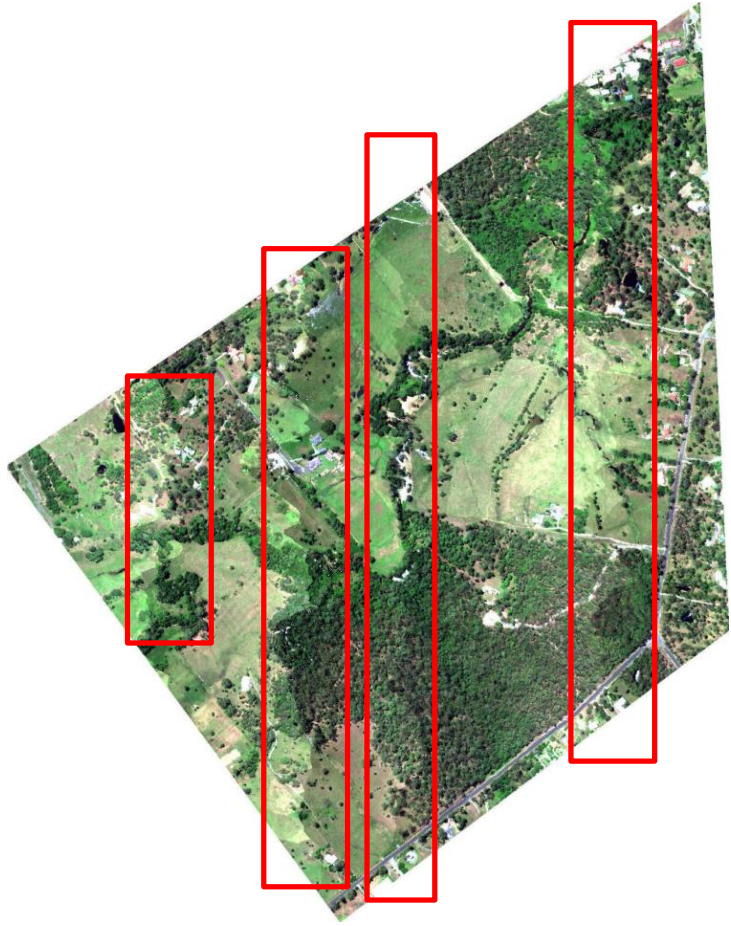

Fig. 3: Examples of the intensity differences across the mosaicked lines.

Based on studies conducted in [28], in order to produce attribute profiles (APs) and their modifications such as ESDAPs, KPCA is able to produce efficient base images in terms of classification accuracies in comparison with other supervised and unsupervised feature extraction techniques. Therefore, here we only investigate KPCA in order to produce base images for ESDAP.

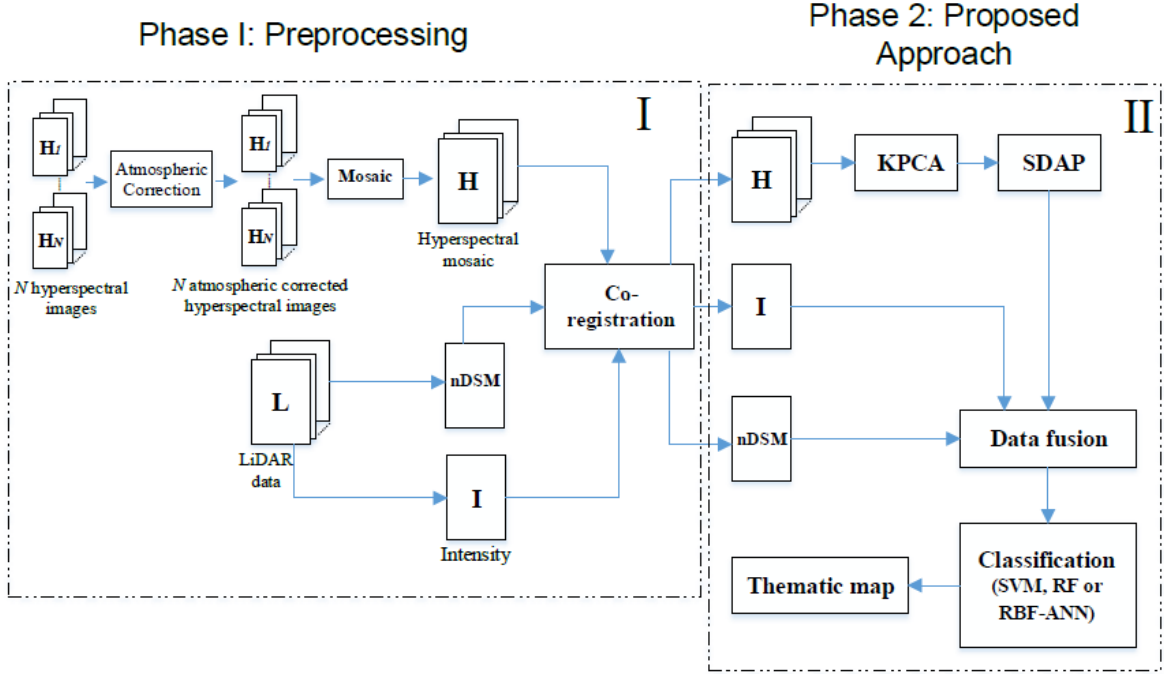

Fig. 4: The general of work flow of the approach developed.

## A. Self-Dual Attribute Profile (SDAP)

Recent efforts in the literature [29, 30] have demonstrated that hyperspectral image classification can greatly benefit from an integrated framework in which both spatial-spectral information are included into the analysis process. In this context, recently region-based filtering tools [31] (called *connected operators*) have received significant attention due to their effectiveness in both extracting spatial information and preserving the geometrical characteristics of the objects in images (i.e., borders of regions are not distorted since only an image is processed by merging its flat zones). Attribute filters [32] are a set of connected operators that are able to simplify a grayscale image according to an arbitrary measure (i.e., attribute).

Dalla Mura *et al.* [21], proposed self-dual attribute profiles (SDAPs) as a variant of APs [33] for the classification of very high geometrical resolution images. The SDAPs are APs built on an inclusion tree [34] instead of a min-tree and max-tree [35], providing a self-dual connected operators $\rho^T$ for a multilevel filtering of both the bright and dark components of an image. SDAPs are obtained by filtering a given grayscale image $u$ with attribute operators using a predicate with increasing threshold values:

$$(1)$$

$$SDAP(u) = \{u, \rho^{T_{A_1}}(u), ..., \rho^{T_{A_{L-1}}}(u), \rho^{T_{A_L}}(u)\}$$

with $\rho$ being the self-dual operator based on the predicate $T$, and $T_\lambda$ a set of $L$ ordered predicates.

In [36] the extended self-dual attribute profiles (ESDAPs) were proposed as the application of SDAPs to hyperspectral data. An ESDAP is obtained by concatenating the SDAPs (i.e., based on one or more attributes) built on one of the $k$ features components $FC_k$ extracted by a feature reduction transformation (e.g., KPCA) from a hyperspectral image:

$$ESDAP = \{SDAP(FC_1), SDAP(FC_2), ..., SDAP(FC_k)\} \qquad (2)$$

The authors in [36] proved that for hyperspectral classification, when both spatial-spectral information are included into the process, ESDAP performs better than EAP [37] (APs extended to hyperspectral images) in terms of classification accuracies. For more information regarding AP and all its modifications, we refer to [28, 33, 38].

*B. Fusion of extracted features via vector stacking*

In order to produce the final classification map, the features produced by the ESDAP are concatenated into a stacked vector along with nDSM and the intensity image. Then the output is classified by SVM, RF or RBFNN. In more detail, let $\zeta_\phi$ be the set of features produced by the ESDAP. The final classification map is obtained by performing one of the aforementioned classifiers on the stacked vector; $\zeta = [\zeta_\phi, \mathbf{nDSM}, \mathbf{I}]^T$.

*C. Experimental Design*

In order to evaluate the efficiency of the proposed method, three different scenarios have been considered to conduct the experiments as follows:

1) *Scenario 1*: In this scenario, the effectiveness of a few strong classifiers (SVM, RF and RBFNN) has been investigated for the classification of the hyperspectral data alone with and without performing KPCA. The output of this scenario infers the effectiveness of the aforementioned classifiers in terms of handling high dimensional data with limited number of training samples as well as the capability of KPCA in order to concentrate almost the whole hyperspectral data in a few features.
2) *Scenario 2*: In this scenario, the importance of the LiDAR derived data (Intensity and nDSM) is investigated as complementary data for improving the classification accuracy of the hyperspectral data and the output of KPCA. The output of this scenario infers the usefulness of LiDAR and hyperspectral data for improving the classification accuracy obtained by SVM, RF and RBFNN.

3) *Scenario 3*: In this scenario, the usefulness of the proposed approach is evaluated by extracting spatial information by ESDAP from both hyperspectral and intensity image. The output of this step infers the effectiveness of considering spatial information extracted by ESDAP along with spectral information extracted by SVM, RF or RBFNN.

To make the classification step more challenging and check the usefulness of the proposed approach for handling volumetric and high dimensional data in an ill-posed situation, we only chose 1% of the reference samples by setting the minimum number of samples to 20 as training and used the rest as test samples. Therefore, our data set contains six classes as follows:

- Bare Ground (20 training samples and 270 test samples),
- Roof-top (22 training samples and 2144 test samples),
- Grass (42 training samples and 4144 test samples),
- Roads (20 training samples and 1088 test samples),
- Trees (29 training samples and 2898 test samples) and
- Water (20 training samples and 818 test samples).

In order to avoid any bias induced by random sampling of the training and test samples, 10 independent Monte Carlo runs are performed and the classification accuracies [Kappa Coefficient (K), Average Accuracy (AA) and Overall accuracy (OA)] and the consecutive CPU processing time are averaged over the 10 runs.

The terms *Hyper*, *nDSM* and *I* refer to situations when the input hyperspectral data, the input nDSM or the input intensity image is classified by the classifiers, respectively. The term *KPCA* is regarded as a situation when KPCA is carried out on the input hyperspectral data and the first informative features with cumulative variance of at least 95% are kept and classified.

In the case of SVM, the Gaussian RBF kernel was applied and the hyperplane parameters have been adjusted by using 5-fold cross-validation. In the case of RF, the classifier has been applied by running down 200 trees. In the case of RBFNN, the Gaussian basis function was adopted. The selection of the centers of the kernel functions was performed by a clustering process (i.e., K-Means algorithm [39]) on the training set by taking into account the class-memberships of the training samples in order to avoid the generation of mixed clusters. The width of a given kernel function was chosen as the standard deviation computed over all training samples included in the cluster associated with the kernel function considered. The weights corresponding to the connections between the hidden and the output neurons were computed by minimizing the sum-of-squares error [40].

April 8, 2016                                                                                                                                                         DRAFT

The adopted kernel function for KPCA is the Gaussian kernel and the parameter $\gamma$ is estimated as the mean value of the distance between each of the samples. The kernel matrix is computed by randomly selecting 500 samples from the total number of pixels present in the image (i.e., in order to perform the transformation in a reasonable processing time).

The ESDAPs were computed by considering the area and the standard deviation attributes. The area allows for the extraction of objects based on their size, while the standard deviation can model the homogeneity of the pixel gray levels belonging to different regions. The selection of the threshold values was automatically performed with the method proposed in [41].

In order to fully exploit the spatial information via ESDAP, one needs to normalize the output of KPCA. The normalization itself helps ESDAP to provide more meaningful features. However, in order to stack LiDAR and hyperspectral data, we have not performed any normalization as they are different types of data and normalizing them in a specific range destroys the information which can be extracted from any of them. Different types of data can provide different information which is helpful to specify class boundaries in the feature space more accurately.

All the experiments were performed in MATLAB on a computer having Intel(R) Core (TM) i7-4710HQ CPU 2.50 GHz and 16 GB of memory.

*D. Experimental analysis and discussion*

*1) Scenario 1:* This experiment analyzes the effectiveness of three well-known classifiers: SVM, RF and RBFNN in two situations: *i)* when the aforementioned classifiers are applied on only the hyperspectral data, and *ii)* when the KPCA is carried out on the hyperspectral data and the informative features are kept with respect to its cumulative variance of at least 95%, and then the features are classified.

As Table I shows, SVM outperforms the other classifiers when the whole hyperspectral data are considered as the input. In this way, SVM improves the OA of RF and RBFNN by almost 3.6% and 5.2%, respectively. In general, SVM is expected to be highly robust in terms of class discrimination when the input data define a highly non-linear problem. In general, RF classifier is based on a tree of weak classifiers, which can take advantage of a large set of redundant features and it can effectively handle noisy data with corrupted bands in a situation when the output of SVM is downgraded due to the quality of the input data. However, for our data set, the noisy and corrupted bands were discarded in the preprocessing step. This may be helpful for SVM to provide better results in terms of classification accuracy. Finally, the RBFNN classifiers are not effective when dealing with a high number of spectral bands, since they are highly sensitive to the Hughes phenomenon. Moreover, the classification result made by RBFNN strongly depends on the selection of the centers and widths of the kernel functions associated with the hidden neurons.

TABLE I: Scenario 1 - Classification accuracies of the proposed methods obtained by different classifiers (AA and OA are reported in percentage and kappa is of no units). The number of features used for classification purposes are reported in the parentheses.

|  |  | SVM | RF | RBFNN |
|---|---|---|---|---|
| Hyper (215) | k | **0.9642±0.0117** | 0.9153±0.0172 | 0.8953±0.0397 |
|  | OA | **97.31±0.878** | 93.63±1.29 | 92.12±3.01 |
|  | AA | **97.69±0.872** | 95.06±1.36 | 92.44±2.77 |
| KPCA (3) | k | **0.9445±0.0059** | 0.9281±0.0115 | 0.8861±0.0264 |
|  | OA | **95.83±0.447** | 94.59±0.874 | 91.42±1.98 |
|  | AA | **96.49±0.587** | 95.42±0.916 | 91.46±2.96 |

TABLE II: Scenario 1: CPU processing time (in seconds) of different classifiers (includes training and test).

|  |  | SVM | RF | RBFNN |
|---|---|---|---|---|
| Hyper (215) | Training | 2.6±0.6 | 0.2±0.01 | 42.3±3 |
|  | Test | 84.9±22 | 78.5±22.6 | 273.7±32.9 |
| KPCA (3) | Training | 0.4±0.0749 | 0.05±0.0254 | 9.9±1.1 |
|  | Test | 5.9±1.4 | 12.2±1.1 | 84.3±3.1 |

The use of KPCA downgrades the OA of SVM and RBFNN by 1.5% and 0.7%, respectively, but improves the OA of RF by 1%. KPCA is an unsupervised feature extraction technique, which is mostly taken into account for data representation into a small number of features. This technique discards class information and just transforms the data into a lower dimensional subspace which is optimal in terms of sum-of-squared error without considering information provided by training samples. Therefore, the use of KPCA does not necessarily lead to better class discrimination and classification accuracies. However KPCA highly decreases the computing time for further classification steps. Table VII shows the CPU processing time (in seconds) for KPCA.

With regard to the standard deviation values reported in Table I, it is interesting to note that SVM provides the most stable results (0.878 for *Hyper* and 0.447 for *KPCA*) over 10 independent Monte Carlo runs. In contrast, RBFNN provides unstable results (3.01 for *Hyper* and 1.98 for *KPCA*) compared to RF and SVM. A classifier is stable if its corresponding classification performance is not highly influenced by making small changes in the training set. The unstable classifiers are characterized by a high variance over a number of independent Monte Carlo runs. On the contrary, stable classifiers have a low variance, but they can have a high bias.

CPU processing time for both test and training are provided in Table II). When we deal with the full dimensional hyperspectral data, RF provides the results in the fastest way. The reason why RF outperforms SVM in terms of CPU processing time is that, while both methods are considered to be effective when dealing with non-linear classification problems, SVM requires a computationally demanding parameter tuning step (cross-validation) in order to obtain optimal results, whereas RF does not require such a tuning process. However, when dimensionality increases using KPCA, SVM can lead to the fastest CPU processing time on test samples. It can be inferred that the CPU processing time of SVM is highly sensitive to the number of bands. Table VII shows the CPU processing time (in seconds) for *KPCA*.

2) *Scenario 2:* In this scenario, we evaluate the usefulness of the nDSM and intensity (I) image along with the hyperspectral data with or without considering KPCA. As the first scenario, the data are classified with SVM, RF and RBFNN.

As reported in Table III, considering the intensity and nDSM as complementary data leads to higher values of Kappa coefficient in all cases compared to a situation when only the hyperspectral data are taken into account. In general, although hyperspectral data can provide detailed spectral information of a scene, the structure and roughness of the surface may influence the spectral responses that reflected from the materials. Moreover, intra-class variability can make spectral responses from the same material even more complicated. Therefore, the consideration of additional information provided by the nDSM and intensity image can lead to better results in terms of classification accuracies. It is interesting to note that the increase in the classification accuracy obtained by the LiDAR data as complementary information is mainly because of the nDSM, while the improvements observed after considering the intensity image are not very significant.

When the whole hyperspectral data are taken into account along with the LiDAR data, SVM significantly outperforms the other two and also provides the most stable results.

When the output of KPCA is taken into account along with the LiDAR data, the results of SVM and RF are comparable and they significantly outperform RBFNN.

Table IV provides the CPU processing time measured for the different approaches studied in Scenario 2. Again, RBFNN outperforms SVM and RF in terms of CPU processing time.

TABLE III: Scenario 2 - Classification accuracies of the proposed methods obtained by different classifiers (AA and OA are reported in percentage and kappa is of no units). The number of features used for classification purposes reported in the parentheses.

|  |  | SVM | RF | RBFNN |
|---|---|---|---|---|
| Hyper + I (216) | k | **0.9765±0.0061** | 0.9213±0.0202 | 0.9375±0.0177 |
|  | OA | **98.23±0.455** | 94.08±1.52 | 95.31±1.33 |
|  | AA | **98.17±0.616** | 95.31±1.55 | 95.15±1.65 |
| Hyper + nDSM (216) | k | **0.9818±0.0073** | 0.9284±0.0021 | 0.9288±0.0019 |
|  | OA | **98.63±0.55** | 94.62±1.57 | 94.65±1.48 |
|  | AA | **98.29±0.643** | 95.64±1.54 | 94.27±1.59 |
| Hyper + I + nDSM (217) | k | **0.9828±0.007** | 0.9318±0.0204 | 0.9435±0.0175 |
|  | OA | **98.71±0.58** | 94.87±1.53 | 95.75±1.32 |
|  | AA | **98.48±0.535** | 95.84±1.32 | 95.55±1.28 |
| KPCA + I (4) | k | **0.9714±0.0044** | 0.9664±0.0121 | 0.9381±0.0194 |
|  | OA | **97.85±0.333** | 97.47±0.911 | 95.34±1.46 |
|  | AA | **97.55±0.668** | 97.09±1.24 | 94.05±1.88 |
| KPCA + nDSM (4) | k | **0.9833±0.0041** | 0.9727±0.0101 | 0.9488±0.0178 |
|  | OA | **98.74±0.312** | 97.94±0.758 | 96.16±1.33 |
|  | AA | **98.29±0.465** | 97.67±0.878 | 93.89±3.1 |
| KPCA + I + nDSM (5) | k | **0.9841±0.0064** | 0.9763±0.0099 | 0.9548±0.0023 |
|  | OA | **98.81±0.482** | 98.21±0.75 | 96.61±1.75 |
|  | AA | **98.38±0.53** | 97.94±0.855 | 94.74±2.37 |

TABLE IV: Scenario 2 - CPU processing time (in seconds) of different classifiers (includes training and test)

|  |  | SVM | RF | RBFNN |
|---|---|---|---|---|
| Hyper + I (216) | Training | 3.3±0.3 | 0.3±0.3 | 41.1±2.3 |
|  | Test | 84.8±15.1 | 115.7±22.8 | 276.9±29.9 |
| Hyper + nDSM (216) | Training | 3.2±0.5 | 0.3±0.08 | 43.1±3.7 |
|  | Test | 84.3±12.8 | 71.8±8.7 | 280.8±33.6 |
| Hyper + I + nDSM (217) | Training | 3.7±0.5 | 0.3±0.07 | 48.6±1.6 |
|  | Test | 87.5±20.3 | 158.6±22.6 | 286.5±34.6 |
| KPCA + I (4) | Training | 0.4±0.04 | 0.1±0.001 | 11.7±1.1 |
|  | Test | 6.6±1.1 | 10.6±1.1 | 84.3±5 |
| KPCA + nDSM (4) | Training | 0.4±0.01 | 0.02±0.001 | 11.5±0.9 |
|  | Test | 6.7±1.1 | 10.3±0.7 | 82.9±3.9 |
| KPCA + I + nDSM (5) | Training | 0.5±0.01 | 0.02±0.002 | 12.5±0.9 |
|  | Test | 7.3±1.6 | 10.9±0.7 | 84.5±4.4 |

TABLE V: Scenario 3 - Classification accuracies of the proposed methods obtained by different classifiers (AA and OA are reported in percentage and kappa is of no units). The number of features used for classification purposes are reported in the parentheses.

|  |  | SVM | RF | RBFNN |
|---|---|---|---|---|
| ESDAP(KPCA) (15) | k | 0.9921±0.0093 | **0.9938±0.0055** | 0.9016±0.0031 |
|  | OA | 99.39±0.697 | **99.54±0.417** | 92.58±2.3 |
|  | AA | 99.07±1.42 | **99.33±0.778** | 91.46±3.49 |
| SDAP(KPCA) + I (16) | k | 0.9896±0.0097 | **0.9939±0.0053** | 0.9303±0.0027 |
|  | OA | 99.22±0.727 | **99.54±0.401** | 94.77±2.09 |
|  | AA | 98.83±1.58 | **99.34±0.747** | 93.58±2.98 |
| SDAP(KPCA) + I + nDSM (17) | k | 0.9904±0.0015 | **0.9957±0.0028** | 0.9277±0.0034 |
|  | OA | 99.28±1.12 | **99.68±0.215** | 94.56±2.64 |
|  | AA | 98.98±1.88 | **99.57±0.285** | 93.87±3.17 |

TABLE VI: Scenario 3 - CPU processing time (in seconds) of different classifiers (includes training and test)

|  |  | SVM | RF | RBFNN |
|---|---|---|---|---|
| ESDAP (KPCA) (15) | Training | 0.8±0.07 | 0.04±0.003 | 19±1.6 |
|  | Test | 15.1±2.2 | 12.8±0.8 | 90.6±4.5 |
| ESDAP (KPCA) + I (16) | Training | 0.9±0.08 | 0.04±0.01 | 18.4±0.9 |
|  | Test | 14.2±2.2 | 12.2±0.6 | 90.3±5.6 |
| ESDAP (KPCA) + I + nDSM (17) | Training | 0.9±0.08 | 0.04±0.002 | 18.2±0.9 |
|  | Test | 14.3±2.3 | 12.5±0.2 | 91.7±6.6 |

3) *Scenario 3:* In this scenario, the importance of considering spatial information in the classification framework is considered. In more detail, the features extracted by KPCA with respect to the cumulative variance of at least 95% are considered as base images to extract spatial information by SDAP. The final classification map is produced by applying SVM, RF or RBFNN.

Although it is possible to apply SDAP on LiDAR data, the consideration of SDAP on LiDAR data can increase the CPU processing time. In addition, SDAP on LiDAR data sometimes leads to the creation of redundant features which downgrade the quality of the classification map. This is the main reason that the ESDAP is only applied on the hyperspectral data.

TABLE VII: CPU processing time (in seconds) for KPCA and ESDAP.

| *KPCA* | ESDAP(KPCA) |
|---|---|
| 111±13 | 112±17 |

By comparing Table V with the previous ones, one can note that when only spectral information derived by hyperspectral, LiDAR or the integration of them is used, the result of the classification (by all the classifiers) is different from the situation when ESDAP is taken into account. This difference indicates that the classification by considering ESDAPs do not necessarily follow the trend of classification with spectral information alone. In this case, the results of SVM and RF are comparable in terms of classification accuracies. The improvement obtained by ESDAP is due to the fact that the application of self-dual connected operators led to an image simplification characterized by more homogeneous regions with respect to the results obtained by extensive or anti-extensive connected operators. Table VII shows the CPU processing time (in seconds) for different preprocessing methods including *KPCA* and ESDAP(KPCA).

Fig. 5 illustrates some of the obtained classification maps for both the worst and best classification overall accuracies reported for the different classifiers for a single training and test set: (a) SVM with *KPCA*, (b) RF with *Hyper*, (c) RBFNN with *KPCA*, (d) SVM with *ESDAP(KPCA)*, (e) RF with *ESDAP(KPCA) + I + nDSM*, (f) RBFNN with *ESDAP(KPCA)*. It should be noted that the intensity changes across the mosaiced lines dramatically influence on the quality of the classification map. As can be seen, when the spatial information extracted by ESDAP is taken into account, the output classification maps of the aforementioned classifiers are more homogeneous with less salt and pepper effects.

In this work, we decided to keep the number of training samples very low on purpose in order to illustrate the performance of the proposed classification techniques in this particular case, which is common in remote sensing applications due to the cost and effort involved in the collection of training data in laborious ground campaigns.

Grass and trees have almost the same spectral signature. In particular, when a feature extraction approach (such as KPCA) is performed on hyperspectral data sets, it gets difficult for the classification approach to discriminate the grass and trees in a proper way since the feature extraction approach does not consider class specific information provided by training samples. On the other hand, LiDAR data can be useful to discriminate trees and grass due to their height difference. However, in this work, we produced 30 features using ESDAP and we have only one feature, i.e., nDSM which considers height differences. When we concatenate them into one stack vector, it means that we put the great emphasis on the spatial information in the hyperspectral data. As a result, the new methodology cannot be very efficient

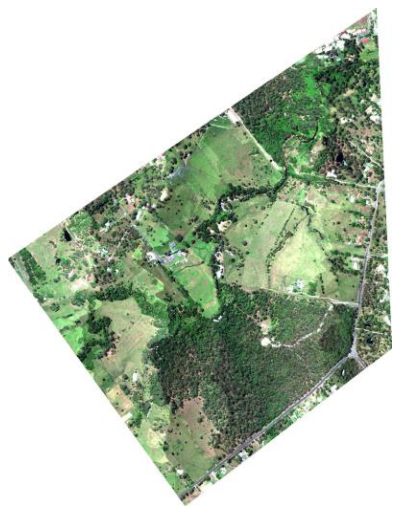

Thematic classes:
- Bare Ground
- Roof-top
- Grass
- Roads
- Trees
- Water

(a)

SVM | RF | RBFNN

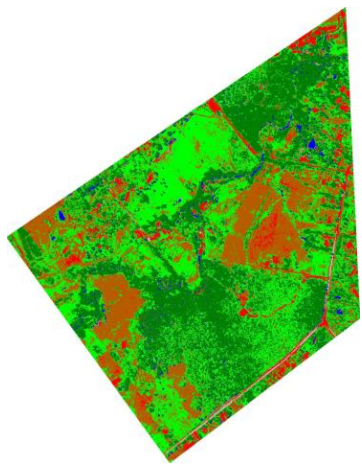 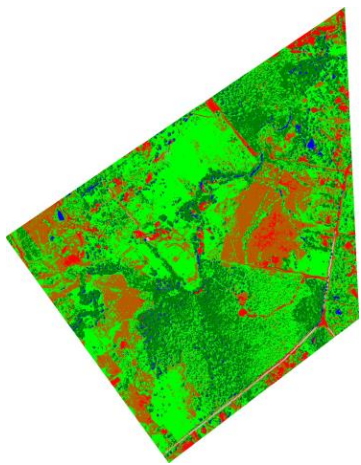 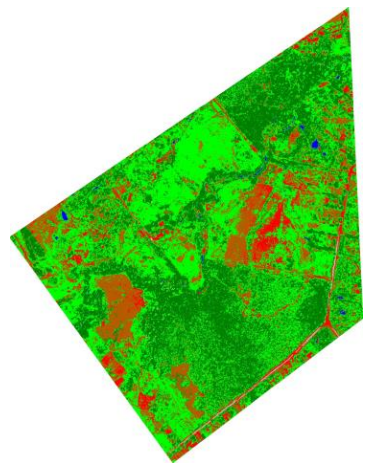

(b) Bare Ground 95.19%, Roof-top 97.95%, Grass 98.36%, Roads 89.80%, Trees 93.17%, Water 99.88%

(c) Bare Ground 82.59%, Roof-top 98.83%, Grass 96.09%, Roads 93.84%, Trees 79.43%, Water 99.88%

(d) Bare Ground 88.15%, Roof-top 79.43%, Grass 96.07%, Roads 71.42%, Trees 92.93%, Water 98.66%

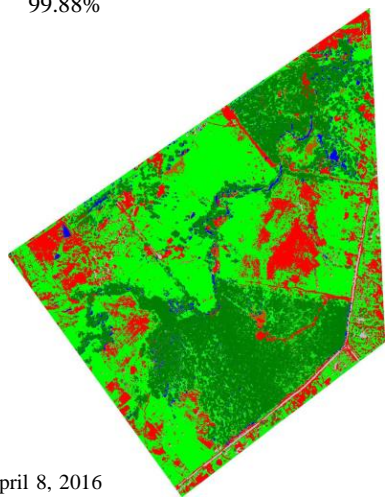 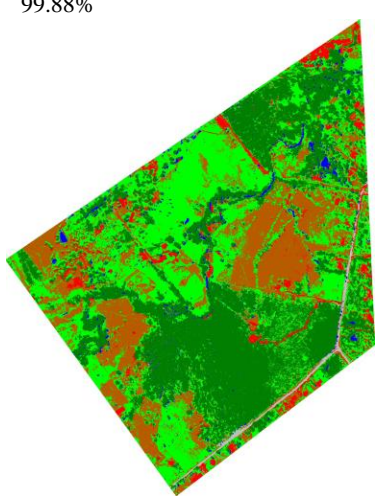 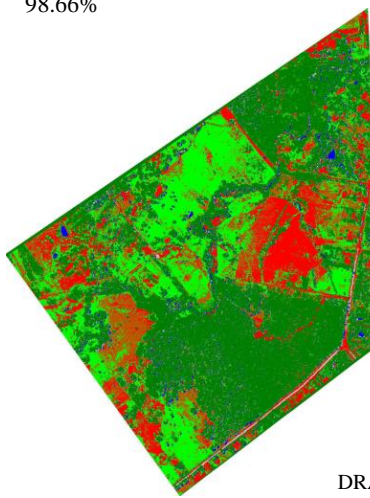

(e) Bare Ground 100%, Roof-top 99.95%, Grass 100%, Roads 99.45%, Trees 99.14%, Water

(f) Bare Ground 98.89%, Roof-top 99.86%, Grass 99.65%, Roads 86.86%, Trees 99.90%, Water

(g) Bare Ground 97.04%, Roof-top 99.95%, Grass 99.52%, Roads 90.07%, Trees 99.65%, Water

to discriminate grass and trees. The same reason can be traced when the classifiers try to classify the rooftops. In this work, we have used very few training samples for each class to make the classification system as challenging as possible.

## V. Conclusion

In this paper, a classification workflow is presented which jointly considers hyperspectral and LiDAR data for accurate land-cover mapping. In the proposed approach, first, KPCA is applied on the hyperspectral data and the first informative features are retained. Then, ESDAP is taken into account to extract the spatial information contained in the hyperspectral data. Furthermore, the produced features as well as the nDSM and intensity images were concatenated into a stacked vector. In order to explore the capabilities of different classifiers, we have investigated SVM, RF and RBFNN to produce the final classification map. Based on the experiments drawn in this paper, the following main observations can be made:

1) When the number of training samples is limited both SVM and RF can provide acceptable results in terms of classification accuracies in both situations, i.e., with or without considering spatial information. In contrast, RBFNN has demonstrated the worst performance compared to SVM and RF.
2) SVM has provided the most stable results in terms of kappa coefficient over a number of independent Monte Carlo runs while RBFNN has shown the worst performance in terms of stability.
3) The proposed classification approach (phase II in Fig. 3) is fully automatic and there is no need to initialize any parameters for the approach.
4) The use of ESDAP can improve the results of SVM, RF and RBFNN in terms of classification accuracies.
5) The increase in the classification accuracy obtained by the LiDAR data as complementary information is mainly because of the nDSM image whereas the improvement by considering the intensity image itself does not significantly influence the obtained results.

As a conclusion, the proposed approach can accurately classify urban areas including both LiDAR and hyperspectral data even if a very limited number of training samples is available. In addition, the proposed methodology is relatively fast and can classify the input data within a short period of time. The proposed classification system is also fully automatic. Furthermore, the use of SVM along with ESDAP is suggested for the classification of urban areas when the number of training samples is limited since: 1) SVM is able to handle high dimensionality with limited number of training samples, and 2) they build up an efficient approach which is able to classify a high volume of urban data quite effectively from the viewpoint of both classification accuracy and computational performance.

As a possible future work and in order to decrease the confusion between different classes such as grass and trees, one may consider more advanced data fusion approaches or put more emphasis on the LiDAR data.

Another possible future research line which deserves to be investigated is that KPCA is an unsupervised approach and it does not consider class specific information provided by training samples. Therefore, the consideration of supervised feature reduction approaches for the proposed framework can be a valuable research area.

Since the number of training samples is limited, another possible research topic is to investigate the capability of semisupervised approaches for the classification of the data composing of both LiDAR and hyperspectral images.

## VI. ACKNOWLEDGMENT

The authors gratefully acknowledge TERN AusCover and Remote Sensing Centre, Department of Science, Information Technology, Innovation and the Arts, QLD for providing the hyperspectral and LiDAR data, respectively. Airborne LiDAR are from http://www.auscover.org.au/xwiki/bin/view/Product+pages/Airborne+Lidar and the airborne hyperspectral are from http://www.auscover.org.au/xwiki/bin/view/Product+pages/Airborne+Hyperspectral licensed by http://creativecommons.org/licenses/by/3.0/. This dataset was made public by Dr. Pedram Ghamisi from German Aerospace Center (DLR) and Prof. Stuart Phinn from the University of Queensland [42].